\documentclass[conference]{IEEEtran}
\IEEEoverridecommandlockouts
\usepackage{cite}
\usepackage{amsmath,amssymb,amsfonts}
\usepackage{algorithmic}
\usepackage{graphicx}
\usepackage{textcomp}
\usepackage{xcolor}
\usepackage{subfigure}

\usepackage[caption=false,font=normalsize,labelfont=sf,textfont=sf]{subfig}
\def\BibTeX{{\rm B\kern-.05em{\sc i\kern-.025em b}\kern-.08em
    T\kern-.1667em\lower.7ex\hbox{E}\kern-.125emX}}

\begin{document}
\title{Using Scene and Semantic Features for Multi-modal Emotion Recognition\\}

\author{\IEEEauthorblockN{Zhifeng Wang\IEEEauthorrefmark{1},
Ramesh Sankaranarayana\IEEEauthorrefmark{2}}
\IEEEauthorblockA{College of Engineering, Computing and Cybernetics,
Australian National University\\
ACT, Australia\\
Email: \IEEEauthorrefmark{1}zhifeng.wang@anu.edu.au,
\IEEEauthorrefmark{2}ramesh.sankaranarayana@anu.edu.au}}
\maketitle

\begin{abstract}
Automatic emotion recognition is a hot topic with a wide range of applications. Much work has been done in the area of automatic emotion recognition in recent years. The focus has been mainly on using the characteristics of a person such as speech, facial expression and pose for this purpose. However, the processing of scene and semantic features for emotion recognition has had limited exploration. In this paper, we propose to use combined scene and semantic features, along with personal features, for multi-modal emotion recognition. Scene features will describe the environment or context in which the target person is operating. The semantic feature can include objects that are present in the environment, as well as their attributes and relationships with the target person. In addition, we use a modified EmbraceNet to extract features from the images, which is trained to learn both the body and pose features simultaneously. By fusing both body and pose features, the EmbraceNet can improve the accuracy and robustness of the model, particularly when dealing with partially missing data. This is because having both body and pose features provides a more complete representation of the subject in the images, which can help the model to make more accurate predictions even when some parts of body are missing. We demonstrate the efficiency of our method on the benchmark EMOTIC dataset. We report an average precision of 40.39\% across the 26 emotion categories, which is a 5\% improvement over previous approaches.
\end{abstract}


\section{Introduction}
People convey their feelings in daily life, and others respond accordingly. Interpersonal ties between many persons are developed via emotional engagement. In particular, automated emotion evaluation is frequently applied in the fields such as robotics \cite{ref-1}, education \cite{ref-3}, marketing \cite{ref-2}, human-computer interaction \cite{ref-37}. Emotions can be represented mathematically as discrete or continuous points in the continuous space of emotional dimensions. The points of Valence, Arousal, and Dominance in three-dimensional space can be used to represent emotion states in continuous space. We concentrate on identifying emotional state in discrete emotion space in this research. Human emotions such as anger, sadness, happiness, excitement, which will be handled in distinct states.

Early research on emotion identification largely used unimodal techniques. The single model relates to speech, gaits, physiological signals, text, body stance, and face expression. The multimodal emotion recognition techniques are then put out after that. Different combinations of a single model were employed in the multimodal emotion recognition techniques to infer emotion states.

Context also plays a crucial part in comprehending human emotion, even though single model and multiple model can extract a individual's traits and increase the precision of inferring human emotion. According to psychology’s context theory \cite{ref-38}, three key components can be identified in the context: focus level 1 local personal factors (such as face, body and pose),focal Level 2 global situational factors (such as scene and semantic information) and focal level 3 cultural factors (friends or strangers). These characteristics, which are given as follows, were combined in our model: \textbf{First}: Three modalities—the face feature extraction network, the body feature extraction network, and the pose feature extraction network—are included in Feature 1 (Local Personal Features). From the input picture, these three modalities will extract regionally specific individual traits. The networks may obtain better inference and offer crucial supplementary information when face or body parts are only partially visible by integrating pose, body, and face modalities. Our experiments demonstrate the importance of local personal characteristics when scene and semantic information are lacking.
 \textbf{Second}: Scene feature extraction network and semantic feature extraction network are two of the modalities included in Feature 2 (Global Situational Factors).A network for extracting scene characteristics will be used to infer certain scene elements near the agent, such a classroom or kitchen scene.Semantic features are the in-depth interpretations or meanings of certain areas or objects in a picture. Semantic cues that surround the agent  such as the sky, grass, river, etc., are features that characterise the environment in which the expression is being done. Semantic characteristics will be more precise in their details while scene features will be more abstract.The semantic features will provide scene characteristics additional information.The agent's emotional states can be impacted by both of them \cite{ref-38}.
\textbf{Third}:Psychology researchers have shown that the proximity and presence of people affect how we feel.When compared to being with friends or alone, being around a stranger will cause your emotions to feel more intense\cite{ref-39}. When two people share an event, they will feel the same feeling \cite{ref-40}. The closeness and interaction with other agents will be investigated for emotion recognition in Feature 3 (Inter-Agent Interactions Feature).

Our objective is to investigate how scene and semantic variables around the agent might increase the accuracy of an automated system for evaluating emotions.

 \textbf{Main contribution}: We consider using a combination of scene and semantic cues to infer emotion states. Our model receives a picture as its input, and as its output, it produces labels for the classification of numerous emotions. The following three aspects can be used to summarise the contributions in this paper:

\begin{itemize}
  \item 
   We recommend using a combination of scene and semantic characteristics to infer emotion states. A scene feature extraction model will specifically assist in extracting characteristics from certain scenes surrounding the agent, such as the class scene and kitchen scene.The semantic feature extraction model will assist in extracting semantic characteristics from the region surrounding the agent, such as the lake, trees, and sky. The accuracy of the model can be significantly increased by combining scene and semantic information.\\
  
 \item 
 We employ a modified version of EmbraceNet \cite{ref-41}  to fuse body and pose characteristics in order to collect body and pose information, which can minimise performance loss from partial data missing.\\
 
 \item 
Our method performs better than previously published methods in the benchmark datasets, the EMOTIC dataset that was introduced in\cite{ref-5}. An average precision score of 40.39\% is reported on EMOTIC, which has 5\% improvement than previous methods \cite{ref-21}
\end{itemize}

\section{Related Work}
\label{sec:related_work}

In this section, we will give a brief introduction about previous efforts on using unique feature and multiple features for emotion recognition, psychology research on context-aware emotion recognition, and context-aware emotion recognition using machine learning.

\textbf{Using unique feature and multiple features for emotion recognition}. In latest years, there has been a significant amount of research on emotion detection. Many earlier attempts use unique characteristics to interpret human feeling such as face\cite{ref-6},  text \cite{ref-13},  body and pose \cite{ref-9}, speech \cite{ref-16}. The capacity to infer emotions utilising the single feature for emotion identification, however, is constrained. Numerous factors have been shown to be useful in multi-label emotion identification in some research\cite{ref-20}.Numerous multimodal emotion identification techniques have been presented out in latest years. Ronak et al. \cite{ref-22} used context and body modalities to categorise emotions. Mittal et al. \cite{ref-21} used a multimodal CNN to classify emotions based on depth, pose, context, and facial characteristic to improve the accuracy in the EMOTIC datasets. The interaction of the agents was discovered using the depth modalities. The other streams use CNNs \cite{ref-48} to extract body features. These two streams are combined in the fusion component to infer emotion states.

\textbf{Context-aware emotion recognition using machine learning}. Context-aware emotion recognition network architectures have been explored in recent years. Kosti et al. \cite{ref-22} deploy two-stream fusion network architecture for context-aware and face-based modalities for emotion detection, which have shown the benefit of context-aware emotion recognition. In order to benefit from the image's context elements, Filntisis et al. \cite{ref-24} and Mittal et al. \cite{ref-21} extracted context characteristics for emotion identification using context-aware modalities. But Filntisis et al. \cite{ref-24} think of optical flow as the second stream and two RGB spatial streams (context and body) as the first stream. By using average score fusion, the two stream fusion's ultimate outcome was obtained. However, Mittal et al. \cite{ref-21} think of the body and face as the first stream and the context as the second. The authors add a third layer to indicate the closeness of the agents and then combine these three streams using an early fusion approach to infer emotional states. They obtained best results in the basic context-based datasets, the EMOTIC datasets. In addition, It was suggested to utilise the modified VGG16 network to extract scene characteristics for emotion identification and the pre-trained Xception network to extract body features\cite{ref-47} .
\begin{figure*}[!ht]
\centering
  \includegraphics[width=\linewidth]{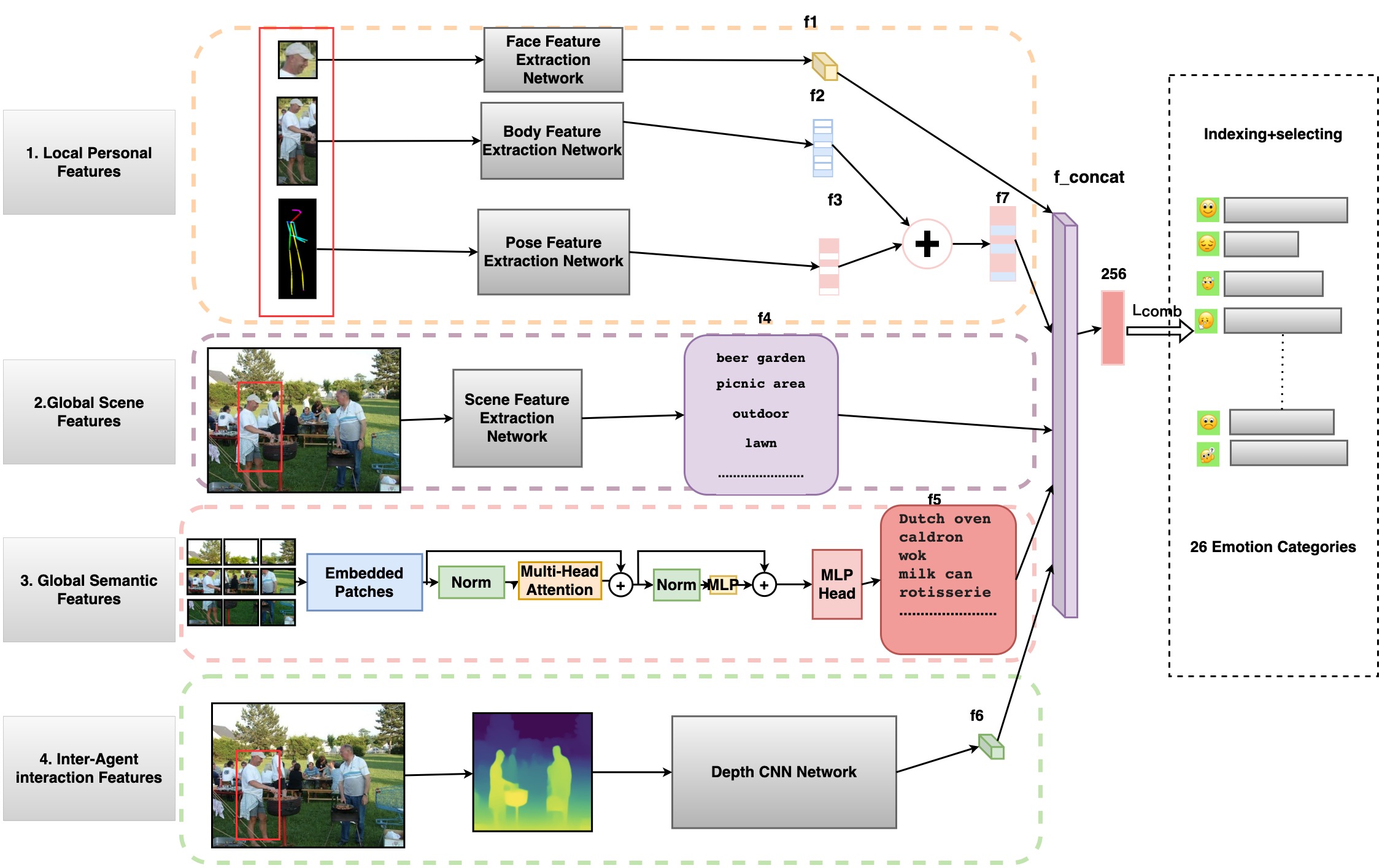}
  \caption{The proposed network architecture for emotion identification. In our model, $f_1$ are extracted face features. $f_7$ are extracted body features. $f_4$ are extracted scene features. $f_5$ are extracted semantic features. $f_6$ are extracted proximity features which are used to calculate distance between target person and other objects. Finally, $f1$, $f7$, $f4$, $f5$, $f6$ will be fused for multi-label emotion identification.}
  \label{network-architecture}
\end{figure*}

\textbf{Attention mechanisms}. With a new attention-based block, Transformers were first presented for machine translation \cite{ref-46}. The attention mechanisms are layers of neural networks that may gather data from the whole input sequence \cite{ref-53}. The primary benefits of attention mechanisms over RNNs on long sequence problems are their technically perfect memory and global calculations. In the disciplines of computer vision, natural language processing, and voice processing, transformers are taking the place of RNNs \cite{ref-54}. 

Transformers were mostly employed in earlier years for natural language processing, and they produced satisfactory accuracy. The benefit of transformers for applications requiring long sequences of data has now been recognised by researchers in computer vision domains. Transformers are now being used in vision - based problems \cite{ref-57}. Dosovitskiy, Alexey, et al.  \cite{ref-29} used the transformer to the picture patches and successfully completed the image classification tasks with the best outcome. For object detection challenges, Carion, Nicolas, et al. \cite{ref-51}  employ the transformer as both encoder and decoder to analyse the connection between the objects and the overall image.

\section{Our Method}
It has both individual traits and features related to the overall context of the image. We suggest utilising a multi-modal network to extract these data, using the diverse features it yields to recognise emotions. Figure  \ref{network-architecture} illustrates the multi-modal network design.RGB picture is the input. In order to create five separate input data for each modality, we first preprocess the RGB image. Face characteristics $f_1$ are extracted using the face feature extraction network. Body and pose characteristics are extracted using the network for extracting body and pose information. In order to merge body and posture features into the combined body features $f_7$, the revised EmbraceNet \cite{ref-41}  will be employed. Our model's improved EmbraceNet can successfully stop performance decline brought by by missing or incomplete data. We suggest using scene and semantic information to infer the target person's emotional states.The network for extracting scene characteristics is responsible for this task -$f_4$. In order to obtain semantic characteristics, we use the semantic feature extraction network-$f_5$. When the target person's face and body traits are only partially visible, the scene and semantic elements can significantly increase the performance of the models. The closeness information $f_6$ between the agents is to be extracted using the depth CNN network. The following step is the concatenation of $f_1$, $f_7$, $f_4$, $f_5$, and $f_6$ for multi-label emotion recognition.

\subsection{Local personal features (Face, Body and Pose) }
Prior research have utilised facial features to determine human emotion \cite{ref-25}. We use MTCNN \cite{ref-26} to get the target subject's facial coordinates. Following that, the facial bounding box is used to crop the target's face. Because the size of the cropped face image is different, we must adjust the face image to fit. The Resnet-101 backbone network is loaded with three channels of input pictures. Research from the previous demonstrates that it is useful for recognising emotions when it comes to body characteristics\cite{ref-25}. In our work, we use a body bounding box to crop the body image from the original picture and then scale it to the same size. Three channels are used as the input for body imaging and are loaded into the Resnet-101 backbone network. In order to obtain the pose coordinates, we utilise Openpose\cite{ref-43}. We then input the coordinates into the modified STGCN \cite{ref-44} model. The body and pose characteristics will then be combined using the improved EmbraceNet, which can reduce performance deterioration caused by missing or incomplete data.

\subsection{Global context features (Scene and Context)}
To reason the emotion states, we suggest combining scene and semantic information. The scene feature extraction network will specifically assist in extracting certain scene characteristics near the agent, such as the class scene and kitchen scene. The network for semantic feature extraction will assist in extracting semantic features surrounding the agent, such as the river, grass, sky, and so on. To extract scene characteristics for our model, we employed a pre-trained CNN network. We use the Resnet-18 as backbone network for scene characteristics extraction. We utilise a modified transformer infrastructure  \cite{ref-29} to capture the semantic characteristics. Multi-head attention, embedded patches, and normalisation are all parts of the transformer design. The use of attention strategies in this section will demonstrate their efficacy in boosting accuracy for tasks like object identification and other ones. The transformer network can assist our model in focusing on some key semantic elements for determining the emotional states of the target individual and giving these key semantic aspects greater weight in our emotion identification tasks. These important semantic features which is helpful for inferring emotion states of target person can be shown in Figure \ref{fig-Qualitative-Results} scene and semantic feature attention map. 1D sequence is sent to the standard transformer. When dealing with 2D pictures $x \in R^{H\times W \times C}$, the 2D pictures must be transformed into 1D patches $x_p\in R^{N \times (P^2C)}$, $C$ is the channel count, $P$ is each picture patch's width and height, N is the number of picture patches overall, $N = HW/P^2$. In the transformer encoder part \cite{ref-46}, It incorporates MLP blocks and multiheaded self-attention (MSA). Prior to each block, layer norm (LN) will be applied, and following each block, residual connections \cite{ref-45}. Two layers of the MLP have a GELU nonlinearity.
\begin{equation}
z_0 = [x_{class}; x_{p}^{1}E;  x_{p}^{2}E;..... x_{p}^{N}E] +E_{pos}
\end{equation}
\begin{align*}
,where E\in{R^{(P^2 C)\times D}}, E_{pos} \in R^{(N+1)\times D}
\end{align*}
\begin{equation}
z_{l}^{1}= MSA (LN(z_{l-1}))+z_{l-1},  l= 1......L
\end{equation}
\begin{equation}
z_{l}= MLP(LN(z_{l}^{1}))+z_{l}^{1},  l= 1......L
\end{equation}
\begin{equation}
y= LN(z_{L}^{0})) 
\end{equation}

\subsection{Inter-Agent Interaction features }
The depth is utilised to represent inter-Agent interaction and it is useful for predicting emotion states, according to earlier study \cite{ref-21}. when there are additional agents all around the agent. They act in a similar way. A person's feelings can be inferred from comparable conduct. To extract interaction characteristics, we noticed a CNN model.Depth maps are the input. The input image $I$ is represented by the $I_depth$. 
\begin{equation}
I_{depth}(i,j) =  d(I(i,j),c)
  \label{depth-equation}
\end{equation}

Where  $d(I(i,j),c)$ indicate the pixel's distance at $ i_{th} $row and $ j_{th} $ column to the camera center $c$. 

\section{Implementation Details}
\subsection{ Data processing}

\textbf{Face and body features}: To extract the target person's facial picture, we employ MTCNN. The body image of the target individual serves as the MTCNN method's input, and its output is the coordinates of the bounding box for the face that has the best chance of being identified. After that the face images will be resized to [128,128] with three channels. Body features are cropped from the entire image using the crop function and the body bounding box coordinates.

\textbf{Depth features}: To obtain the depth maps from the entire picture, we utilise MiDaS \cite{ref-30}. Depth maps with the size [224,224,1] are the output. The depth maps can be calculated by Equation \ref{depth-equation}

\subsection{Network Architecture}

\textbf{Face and body features}:The size of the input picture for the facial features extraction network is [224, 224, 3]. Resnet-101, which was trained on ImageNet and finetuned on EMOTIC dataset, is the fundamental network for face features extraction network. The network structure of the body features extraction network is the same.

\textbf{Context-aware features (Scene and Context)}: The Resnet network, which was previously trained on the place 365 datasets, serves as the backbone of the scene features extraction network. The entire image with three channels is the input to the semantic features extraction network. In the semantic feature extraction network, it includes multi-headed self-attention, layer norm, and MLP block. The structure can be shown in Fig.\ref{network-architecture}. 

\textbf{Depth features}: The CNN depth network uses depth maps as input, which are generated using Equation\ref{depth-equation}. There are five 2D convolutional layers in the CNN network.The image size will be reduced after each convolutional layer, which will be used to obtain fine-grained features.

\textbf{Fusion method}: We apply the early fusion approaches to merge the face, body,   context, pose and depth characteristics. Instead of utilising single features to infer emotion states, the feature vectors for emotion detection are fused, $f_{concat}$ = [$f1$, $f7$, $f4$, $f5$, $f6$]. Two fully connected layers are applied on merged features. The dimension for first fully connected layer are  transfer from  256 to 26.26 are corresponding to 26 emotion categories. The second fully connected layer's dimension transfer from 256 to 3. The softmax layer comes after these  fully connected layer. Loss and error will be computed using the output. 

\begin{table*}[htbp]
\centering
\caption{The EMOTIC dataset contains 26 different kinds of emotions.Each person is capable of having several labels that correlate to various emotion groups. When comparing EMOTIC dataset  \cite{ref-35} to other datasets like AFEW\cite{ref-32},  AffectNet \cite{ref-34},CAER-S  \cite{ref-33} and CAER  \cite{ref-33} , the photos in the EMOTIC collection were taken in the outdoors and contain a large number of context characteristics.}
\begin{tabular}{c c c c c c} 
\hline \hline
 Data type& Dataset & Number of images  & Settings  & Emotion categories  & Context \\ 
\hline
 Images & EMOTIC \cite{ref-35}&  23,571 & Web &  26 Categories & Yes \\ 
\hline
 Images &CAER-S \cite{ref-33} & 70,000  & TV show & 7 Categories & Yes  \\ 
\hline
 images &AffectNet \cite{ref-34} & 450,000  &  Web&  8 Categories & No \\ 
\hline
 Videos &CAER \cite{ref-33} &13,201 clips  &TV show & 7 Categories  & Yes \\ 
\hline
Videos & AFEW \cite{ref-32} & 1,809 clips  &Movie & 7 Categories & No \\
\hline
\end{tabular}
 \label{datasets-compare}
\end{table*}

\begin{figure*}[!h]
\centering
\subfigure[EMOTIC\cite{ref-35}]{
\begin{minipage}[t]{0.3\textwidth}
\centering
\includegraphics[width=\linewidth]{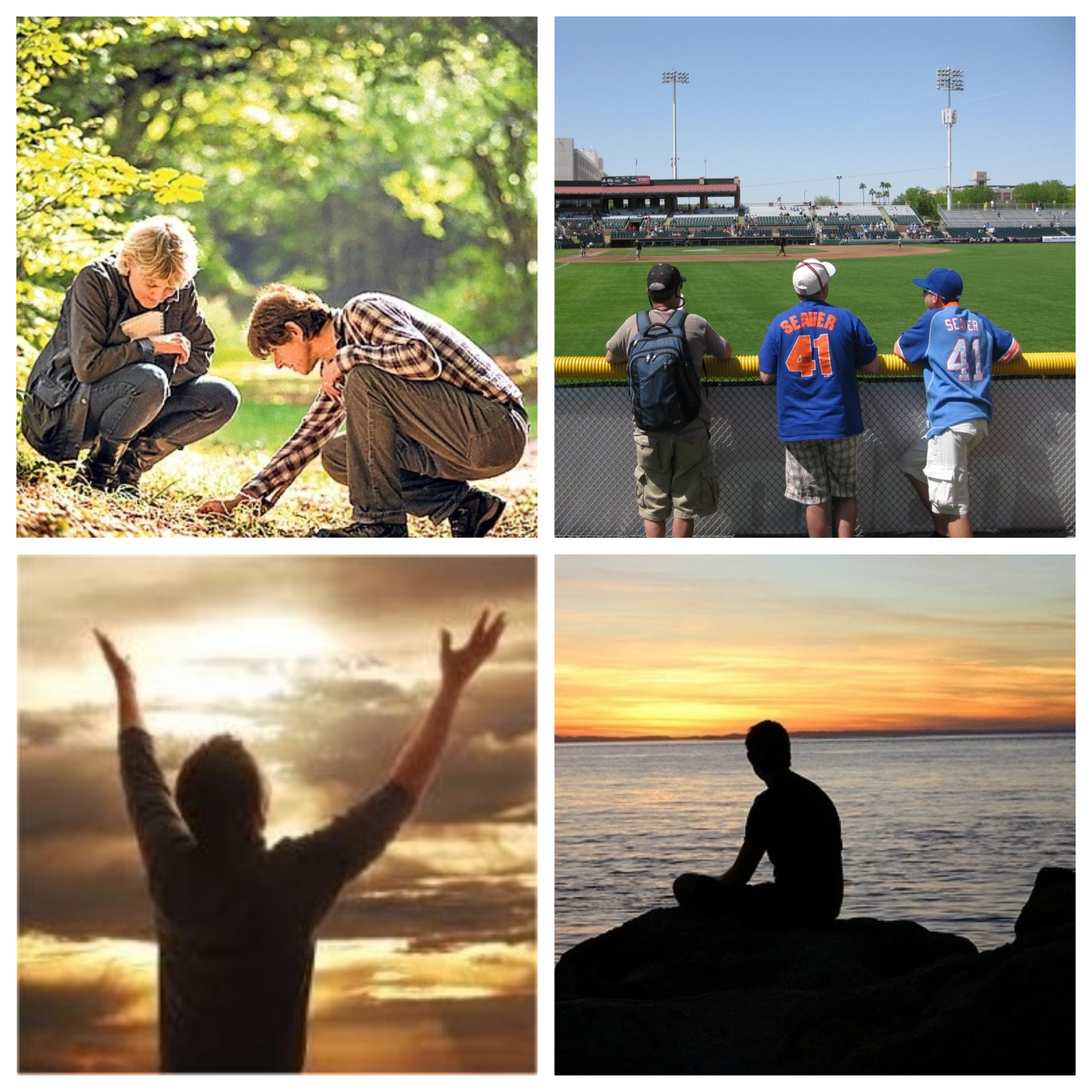}
\end{minipage}
}%
\subfigure[AffectNet \cite{ref-34}]{
\begin{minipage}[t]{0.3\textwidth}
\centering
\includegraphics[width=\linewidth]{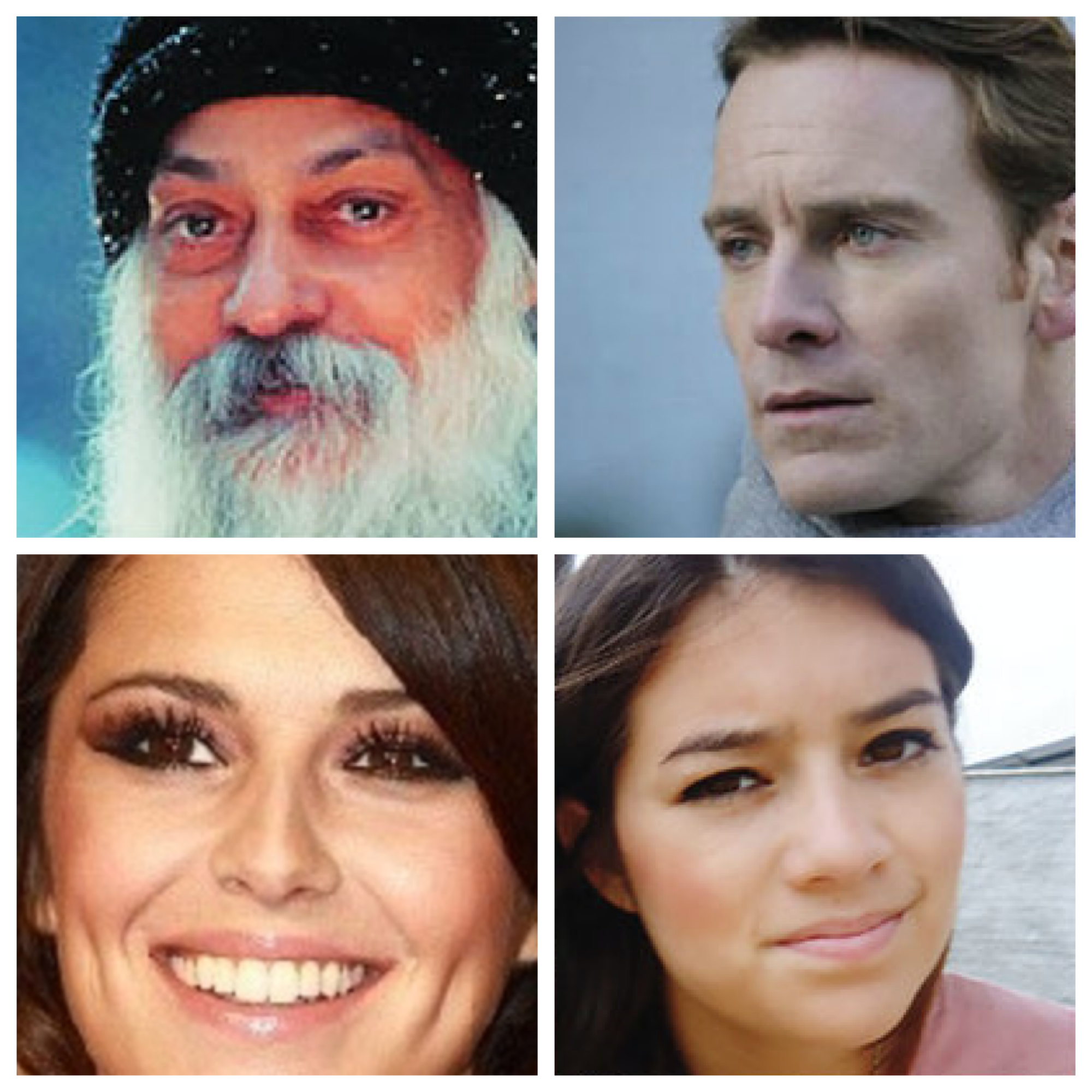}
\end{minipage}%
}%
\subfigure[AFEW \cite{ref-32}]{
\begin{minipage}[t]{0.3\textwidth}
\centering
\includegraphics[width=\linewidth]{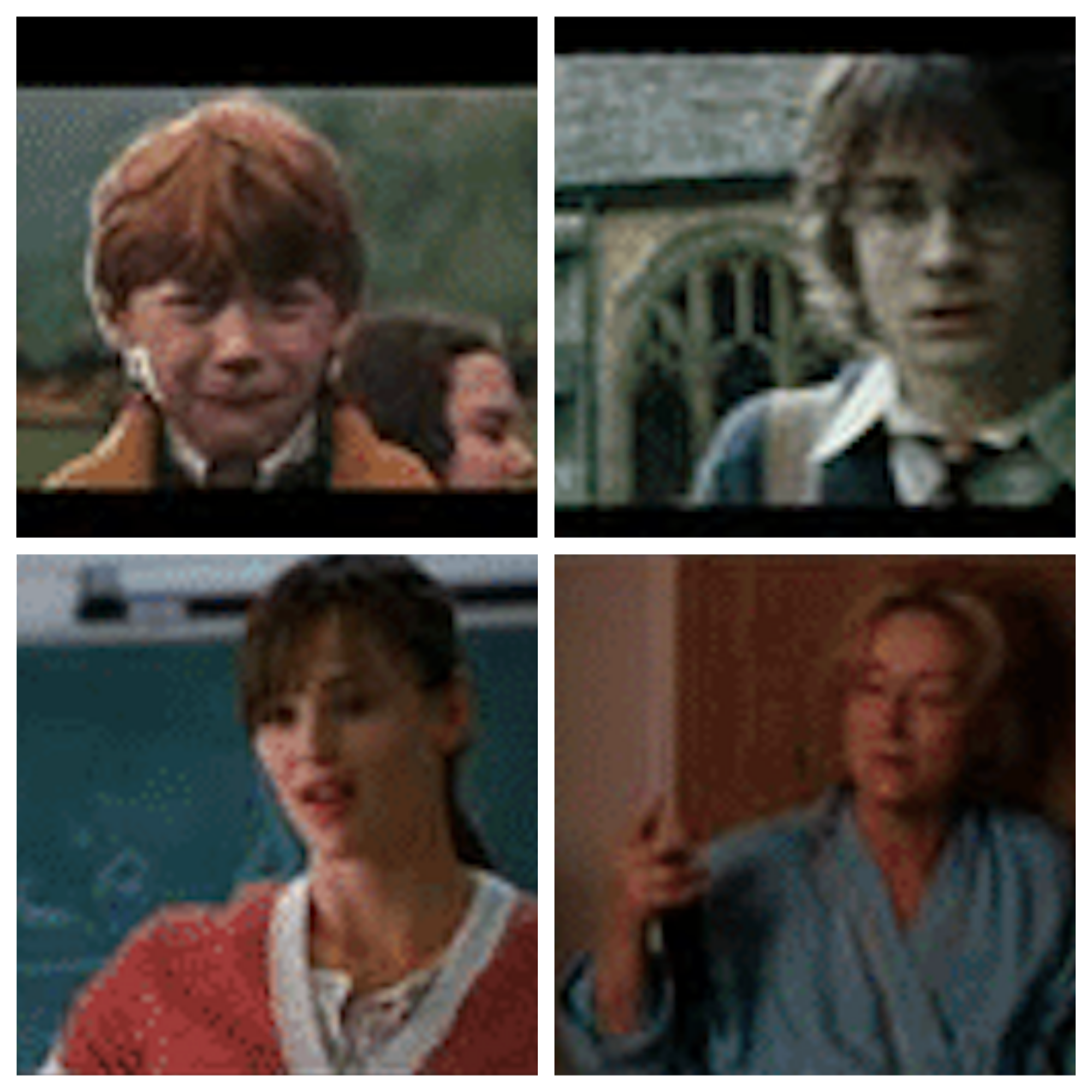}
\end{minipage}
}%
\centering
\caption{The photos in the EMOTIC collection were taken in the outdoors and contain a large number of context characteristics. Only face photos with little background data are present in the AffectNet \cite{ref-34} collection. The AFEW dataset \cite{ref-32} contains pictures taken from movies with partial body and facial information, but little background data.}
 \label{emotic-dataset}
\end{figure*}

\textbf{Loss function}:
A weighted mixture of two independent losses makes up the loss function. The prediction of 26 categories plus the forecast of 3 continuous dimensions make up the prediction $y_{pred}$. $y^{pred} = (y^{disc}, y^{cont})$., where $y^{disc} =(y_{1}^{disc}, y_{2}^{disc}.....,y_{26}^{disc}) $ and $y^{cont} =(y_{1}^{cont}, y_{2}^{cont}, y_{3}^{cont}) $. The definition of the combination loss is \\
\begin{equation}
L_{comb} = \lambda_{disc} L_{disc}+ \lambda_{cont}L_{cont}
\label{L-combination}
\end{equation}
Where the $ \lambda_{disc} $ and $ \lambda_{cont} $ will take each loss's contribution into account.\\
We employ the weighted Euclidean loss for discrete categories in our studies.Compared to Kullback-Leibler loss and multi-class multi-classification hinge loss, the Euclidean loss is more efficient. The following definition applies to discrete categories loss:

\begin{equation}
L_{disc} = \sum_{i=1}^{26}w_{i}(y_{i}^{real disc} - y_{i}^{pred disc})^2
\label{L-discrete}
\end{equation}
where the real label is $y_{i}^{real disc}$ and  the prediction label for the $i_{th}$ emotion category is $y_{i}^{pred disc}$. The weight for every emotion category is  the parameter $w_{i}$ and $w_{i} = \frac{1}{ln(c+p_{i})}$. $p_{i}$ is the probability for $i_{th}$
category and c is to control the range of valid values for weight.\\
It has been demonstrated that the frequently used $L_{cont}$  is less dependent on outliers and is described as follows \cite{ref-31}.
\begin{equation}
L_{cont} = \sum_{k=1}^{3}v_{k} 
\begin{cases}
      0.5x^2 & \mbox{if } |x_k| < 0,\\
      |x_k|-0.5 & \mbox otherwise
  \end{cases}
  \label{L-continuous}
\end{equation}
where  $v_{k}$ is weight given to every continuous emotion category and $x_{k} = (y_{k}^{real cont}- y_{k}^{pred cont})$.
\section{Datasets}
The context-based EMOTIC datasets will be introduced. There are 23,571 photos of 34,320 labelled people in the wild places in the EMOTIC database\cite{ref-35}.The EMOTIC dataset contains 26 different kinds of emotions.Each person is capable of having several labels that correlate to various emotion groups. When comparing EMOTIC dataset  \cite{ref-35} to other datasets like AFEW\cite{ref-32},  AffectNet \cite{ref-34},CAER-S  \cite{ref-33} and CAER  \cite{ref-33} , the photos in the EMOTIC collection were taken in the outdoors and contain a large number of context characteristics. Many of the intended people's faces are hidden, thus we need to develop a network that can successfully extract background data for emotion identification. The EMOTIC dataset has 26 categories for emotions. Only 7-8 emotion categories are included in other datasets.  Table  \ref{datasets-compare} provides an overview of the dataset details. Figure  \ref{emotic-dataset} displays some of the EMOTIC's sample pictures. Only face photos with little background data are present in the AffectNet \cite{ref-34} collection. The AFEW dataset \cite{ref-32} contains pictures taken from movies with partial body and facial information, but little background data.

\section{Experiments and Results}
\subsection{Training details}
We use the foundational context-based emotion datasets from EMOTIC to train our model using stochastic gradient descent with momentum. Our approaches have a batch size of 52 and a 45 epoch. The learning rate starts off at 0.001 and decreases by a factor of 10 every 15 epochs when we employ the Adam optimizer.

\subsection{Ablation Study}

We train our model on the new dataset and EMOTIC dataset. The new dataset contains four fundamental emotion categories, in order to comprehend model's components.

\textbf{Benefits of Local Personal Features.} We conduct an experiment using the EMOTIC dataset and a subset of EMOTIC with four fundamental emotions in order to analyse the effects of local personal variables for predicting emotion states of the target person. In the Table \ref{ablation-experiments-ap}, using local personal features (Feature 1), our model can achieved good results in following emotion categories such as 'Anticipation', 'Confidence', 'Engagement', 'excitement' and 'Happiness'. The reason is that inferring these emotion categories relies on face and body features. In addition, local personal features are basic features for emotion recognition. Retaining local personal features will improve inferring ability of our model. 

\begin{figure*}[htbp]
\centering
\subfigure[Face, body and depth ]{
\begin{minipage}[t]{0.33\linewidth}
\centering
\includegraphics[width=\linewidth]{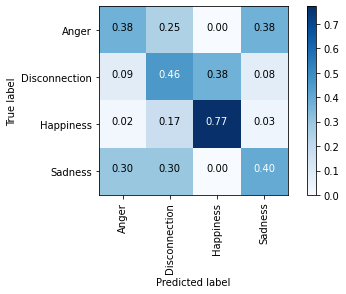}
\end{minipage}
}%
\subfigure[Face, body  and context]{
\begin{minipage}[t]{0.33\linewidth}
\centering
\includegraphics[width=\linewidth]{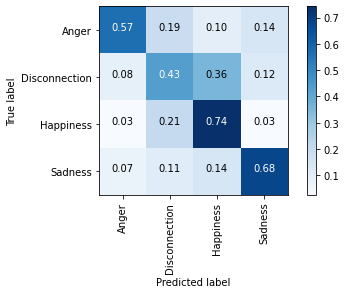}
\end{minipage}%
}%
\subfigure[Face, body, depth  and context]{
\begin{minipage}[t]{0.33\linewidth}
\centering
\includegraphics[width=\linewidth]{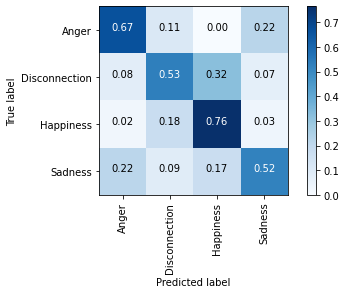}
\end{minipage}
}%
\centering
\caption{Matrix of confusion confirms that, in order to improve model's performance,  multi-modal combination with merged scene and semantic information is required}
 \label{newdata_confusion_matrix}
\end{figure*}
\textbf{Benefits of Global Scene and Semantic Features.} 
Comparing with $2^{nd}$ column and $3^{rd}$ column in the Table \ref{ablation-experiments-ap}(a), it shows that after adding scene and semantic features, the Average Precision Score increase about 6\%, from 33.52\% to 39\%. In the Table \ref{ablation-experiments-ap}(b), the Average Precision Score increase about 13\%, from 42.53\% to 55.23\%.  This is due to that the EMOTIC dataset include rich background information. Our global scene and semantic features extraction network can effectively extract these background information. So, our model with combined scene and semantic features can perform well either in the large number of emotion categories or the small number of emotion categories (only 4). The accuracy of the model on all emotion categories have an improvement. 

\begin{table}[htbp]
 \caption{Ablation Experiments: isolating impact of local features and global Features analysing and interpreting the various elements using the AP score. Feature 1 - Face, body and pose feature. Feature 2 - Scene and semantic features. Feature 3 - Depth features }
 \subfigure[AP Score for 26 emotion categories on EMOTIC Dataset]{
\begin{minipage}[t]{\linewidth}
\centering
\resizebox{\textwidth}{50mm}{
\begin{tabular}{| c | c | c | c | c | } 
\hline
 Labels& Only 1 &  Only 1 and 2 & Only1 and Only 3 & 1, 2 and 3  \\ 
\hline
 Affection & 38.35 &  48.58 &42.74 &49.53  \\ 
\hline
 Anger & 21.87 &  28.79&22.67 &30.46 \\ 
\hline
 Annoyance & 20.99 &23.26  & 25.24  &29.55 \\ 
\hline
 Anticipation &94.95  & 95.74 & 95.57&96.15 \\ 
\hline
 Aversion &16.50  & 21.15 & 19.65 &23.63\\ 
\hline
 Confidence &  76.16& 80.46 &78.90 &81.00  \\ 
\hline
 Disapproval &25.08  &31.71  & 29.11&34.33 \\ 
\hline
 Disconnection &  36.60&  42.78 & 40.05 &46.14   \\ 
\hline
 Disquietment &  16.44&21.50  &21.67 &24.78   \\ 
\hline
 Doubt/Confusion &  19.63 & 20.42 & 25.01&24.96   \\ 
\hline
 Embarrassment &  5.72& 7.17 & 7.21 &7.55  \\ 
\hline
 Engagement &98.12  &98.31  & 98.06& 98.36 \\ 
\hline
 Esteem & 27.02 &28.55  & 27.41 &28.63\\ 
\hline
 Excitement & 76.49 & 81.34 & 80.42&82.92 \\ 
\hline
 Fatigue & 11.65 & 21.96 &12.81  &22.66  \\ 
\hline
 Fear &9.14  & 12.86 & 10.45 &12.18  \\ 
\hline
 Happiness & 79.99 &  84.06& 86.19& 87.56  \\ 
\hline
Pain  & 16.87 & 24.88 & 21.74 &25.56 \\ 
\hline
 Peace & 27.86 & 33.91 &  28.90&33.28\\ 
\hline
 Pleasure &  50.08& 55.68 &58.16 &58.86  \\ 
\hline
 Sadness &  17.58& 36.22 & 25.63&35.82  \\ 
\hline
 Sensitivity & 7.31 &11.99  & 8.70&12.38  \\ 
\hline
 Suffering & 17.09 &36.13  & 21.32 &36.89 \\ 
\hline
 Surprise & 14.92 & 15.90 & 14.43 &14.45   \\ 
\hline
 Sympathy &32.24  &37.50  & 35.29 &38.67   \\ 
\hline
 Yearning &12.76  &13.17  &  12.84&13.72  \\ 
\hline
mAP & 33.52 & 39.00 & 36.57 &40.39\\
\hline
\end{tabular}}
\end{minipage}
}%

\subfigure[AP Score for four basic emotion categories on new  Dataset ]{
\begin{minipage}[t]{\linewidth}
\centering
\resizebox{\textwidth}{12mm}{
\begin{tabular}{| c | c | c | c | c | } 
\hline
 Labels&Only 1& Only 1 and 2 & Only 1 and Only 3 & 1,2 and 3 \\ 
\hline
 Anger &20.23 &36.41  &24.39  & 33.17\\ 
\hline
 Disconnection &  35.89& 42.44 &45.11 & 48.15 \\ 
\hline
 Happiness & 87.75 & 88.87& 92.68 &  93.53\\ 
\hline
 Sadness &26.26 &  53.22& 27.48 &  49.87 \\ 
\hline
mAP & 42.53 & 55.23&47.42 &  56.18 \\
\hline
\end{tabular}}
\end{minipage}
}%
  \label{ablation-experiments-ap}   
\end{table}

 \textbf{Benefits of Depth Features.} Comparing with $2^{nd}$ column and $4^{th}$ column in the Table \ref{ablation-experiments-ap}(b), it shows that after adding depth features, the emotion category 'Disconnection' increase about 10\%, from 35.89\% to 45.11\%. The reason is that depth features can be used to calculate distance between target person and other agents. If the distance between target person and other agents is far away, it indicates that the target person is disconnected with other people. So, the emotion state of target person is 'Disconnection'. Therefore, it verifies that depth feature will boost the recognition performance.

 \begin{table}[!ht]
\begin{minipage}[t]{\linewidth}
\centering
\caption{We analyse our findings using the common Average Precision (AP) metric, and we compare our results with previous results for 26 emotion categories on EMOTIC dataset}
\resizebox{\textwidth}{54mm}{
\begin{tabular}{| c | c | c | c | c | c |  } 
\hline 
 Labels& Kosti et al.  \cite{ref-22} & Zhang et al. \cite{ref-42} & Mittal et al. \cite{ref-21} & Ours ($L_{comb}$)& Ours ($L_{disc}$) \\ 
\hline
 Affection & 27.85 &  46.89 & 45.23 &48.89 & \textbf{49.53} \\ 
\hline
 Anger & 09.49 &  10.87& 15.46 &24.29 & \textbf{30.46}\\ 
\hline
 Annoyance & 14.06 &11.23  &  21.92& 28.42&  \textbf{29.55} \\ 
\hline
 Anticipation &58.64  & 62.64 &72.12  & 96.07& \textbf{96.15} \\ 
\hline
 Aversion &07.48  & 5.93 & 17.81&22.21 & \textbf{23.63}\\ 
\hline
 Confidence &  78.35& 72.49 & 68.65 & 80.78& \textbf{81.00}  \\ 
\hline
 Disapproval &14.97  &11.28  & 19.82 &31.99 & \textbf{34.33} \\ 
\hline
 Disconnection &  21.32&  26.91 & 43.12 &41.63 & \textbf{46.14}  \\ 
\hline
 Disquietment &  16.89&16.94  &18.73  & 23.73&\textbf{24.78} \\ 
\hline
 Doubt/Confusion &  29.63 & 18.68 & \textbf{35.12} & 26.13&24.96  \\ 
\hline
 Embarrassment &  03.18& 1.94 & \textbf{14.37} & 8.23&7.55  \\ 
\hline
 Engagement &87.53  &88.56  & 91.12 & 98.22&\textbf{98.36} \\ 
\hline
 Esteem & 17.73 &13.33  & 23.62 & 27.44&\textbf{28.63} \\ 
\hline
 Excitement & 77.16 & 71.89 & 83.26 & \textbf{83.49} &82.92  \\ 
\hline
 Fatigue & 09.70 & 13.26 & 16.23 &16.79 &\textbf{22.66}  \\ 
\hline
 Fear &14.14  & 4.21 & \textbf{23.65} &11.67 &12.18 \\ 
\hline
 Happiness & 58.26 &  73.26& 74.71 & 87.41& \textbf{87.56} \\ 
\hline
Pain  & 08.94 & 6.52 & 13.21 &24.58 & \textbf{25.56}\\ 
\hline
 Peace & 21.56 & 32.85 & \textbf{34.27} & 30.18&  33.28\\ 
\hline
 Pleasure &  45.46& 57.46 &\textbf{65.53}  & 59.53& 58.86\\ 
\hline
 Sadness &  19.66& 25.52 & 23.41 & 33.19& \textbf{35.82} \\ 
\hline
 Sensitivity & 09.28 &5.99  &8.32  & 10.34 & \textbf{12.38}\\ 
\hline
 Suffering & 18.84 &23.39  & 26.39 & 28.46& \textbf{36.89} \\ 
\hline
 Surprise & \textbf{18.81} & 9.02 & 17.37 &14.01 &14.45  \\ 
\hline
 Sympathy &14.71  &17.53  & 34.28 &38.44 & \textbf{38.67}  \\ 
\hline
 Yearning &08.34  &10.55  & \textbf{14.29} & 13.60 &13.72 \\ 
\hline
mAP & 27.38 & 28.42 & 35.48 & 38.84& \textbf{40.39}  \\
\hline
\end{tabular}}
 \label{results-for-methods}
 \end{minipage}
\end{table}

\begin{table}[!ht]
\centering
\caption{We analyse our findings using the common Average Precision (AP) metric, and we compare our results with previous results for four basic emotion categories on new dataset}
\resizebox{\linewidth}{10mm}{
\begin{tabular}{| c | c | c | c | c | } 
\hline
Labels& Kosti et al.  \cite{ref-22} & Zhang et al. \cite{ref-42} & Mittal et al. \cite{ref-21} &  Ours  \\ 
\hline
 Anger & 21.50& -  &15.85  &   \textbf{33.17}\\ 
\hline
 Disconnection & 37.76 & -  &38.76  & \textbf{48.15}  \\ 
\hline
 Happiness &88.84 & -&85.76  &  \textbf{93.53} \\ 
\hline
 Sadness &30.33 & - &48.29 &  \textbf{49.87} \\ 
\hline
mAP & 44.61 & - & 47.21& \textbf{56.18}  \\
\hline
\end{tabular}
}
\label{results-for-methods-newdataset}
\end{table}
 
 \textbf{Confusion matrix analysis.} Comparing Figure \ref{newdata_confusion_matrix}(a) and Figure \ref{newdata_confusion_matrix}(c), only using face, body and depth features stream, the networks will perform poorly and  the network have limited ability to distinguish 'Anger' and 'Sadness'. After  adding context streams, the joint model will greatly improve the emotion recognition performance on 'Anger', 'Disconnection', 'Sadness'.  The accuracy for 'Anger', 'Disconnection' and 'Sadness' increased by 29\%, 7\% and 12\%. This experiment verifies our concern that we need to use multi-modal combination  with combined scene and semantic features to boost the performance on all emotion categories.
\subsection{Analysis and Discussion}
\textbf{Comparison with SOTA}:  We analyse our findings using the common Average Precision (AP) metric, and we compare our results with SOTA results in Table \ref{results-for-methods}. For EMOTIC Dataset, by using combination loss, our model improves by 3\% and achieve 38.84\% Average Precision (AP) score, outperforming all other methods. Our best model use the discrete categories loss and achieved 40.39\% Average Precision (AP) score, outperforming by a margin of 4.91\% Mittal's method. We extract four basic emotion categories and summarize the AP score in the new dataset in Table \ref{results-for-methods-newdataset}. The earlier approaches Mittal et al. \cite{ref-21}  and Zhang et al. \cite{ref-42}  don't have publicly accessible implementations, therefore we implement their algorithms in accordance with the paper's statement and our best knowledge to evaluate them utilizing the same settings on the new dataset.On a new dataset, our technique outperformed all others, scoring 56.18\% Average Precision (AP) .

\begin{figure}[htbp]
\centering
  \includegraphics[width=\linewidth]{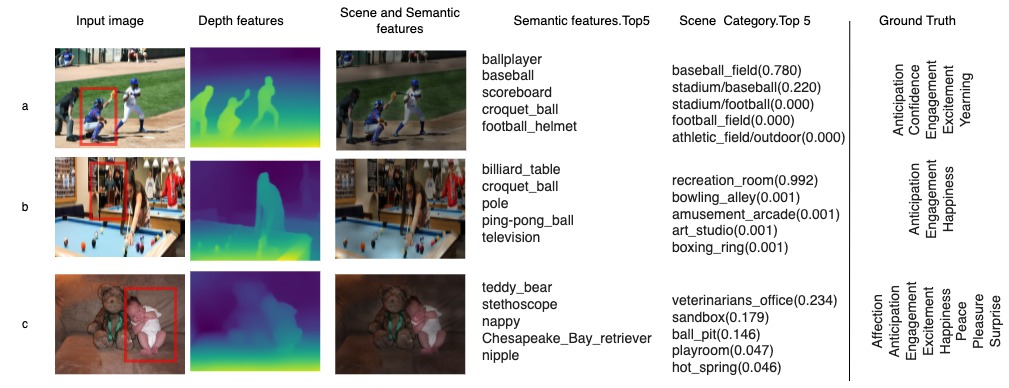}
  \caption{In our model, the scene extraction network can extract scene features and the semantic extraction network can extract semantic features. Our model can effectively distinguish different objects in the same scene.}
  \label{fig-Qualitative-Results}
\end{figure}

The key to our network's efficiency is the integration of several modalities, and we use transformer-based attention modalities in this system to capture background information. For instance, in the Figure \ref{fig-Qualitative-Results}.(a), we can see that the person's face is hidden in the first picture.The network is unable to obtain facial features with acceptable quality. Therefore, the global scene and semantic characteristics are given greater consideration by our network. We can see from the scene elements that the baseball players are on the baseball field playing the game. The semantic features shows that there are 'ball players', 'baseball', 'scoreboard', 'croquet ball' and 'football helmet' in baseball field. Our model utilises the global scene and semantic features to successfully infer emotion states of target person. 

Comparing with Kosti et al.\cite{ref-22}'s method,the context modality can only give scene category information and does not employ attention strategies in their network, despite the fact that they use two CNN streams to extract context features and body features for emotion identification.Their network is unable to properly change its weights and focus its attention to the context's most important elements. In the Table \ref{results-for-methods},  their method only achieved 27.38\% average precision score. 

\begin{figure}[h!]
\centering
  \includegraphics[width=\linewidth]{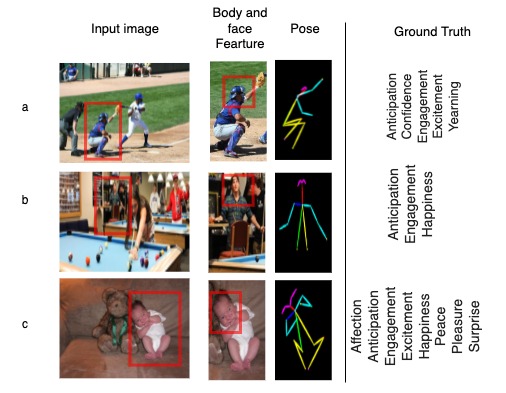}
  \caption{We employ a  fusion network to fuse body and pose characteristics in order to collect body and pose information, which can minimise performance loss from partial body and pose data missing.}
  \label{fig-scene-semantic}
\end{figure}

Comparing with Zhang et al's \cite{ref-42} method, they capture context information using the Region Proposal Network (RPN) and feed the result  to the Graph Convolutional Network (GCN). The other stream uses CNN to gather body characteristics. These two streams are combined in the fusion component to estimate emotion. The CNN network can extract limited body features. But this CNN network can not extract pose features. However, our method can utilise pose features for emotion recognition. The pose features can be shown  in the Figure \ref{fig-scene-semantic}. In addition, region proposal network (RPN) do not utilise self attetion techniques to extract context features for emotion recognition. Their network can not arrange higher weights to important part of context. So their method performs poorly in EMOTIC dataset and new dataset. In the Table \ref{results-for-methods},  their method only achieved 28.42\% average precision score. 

Comparing with Mittal et al. \cite{ref-21} method, they try to use a self-attention-based CNN to extract semantic features, they achieved great improvement on average precision score for emotion recognition. However, their self-attention-based CNN has limited ability to understand what different objects there are in the same scene. In our model, the scene extraction network can extract scene features and the semantic extraction network can extract semantic features. Our model can effectively distinguish different objects in the same scene. For instance, in the Figure \ref{fig-Qualitative-Results}.(b), the scene is recreation room. In this recreation room, our model can get specific semantic features such as billiard table, croquet ball, pole, television and so on. So, our model achieved competitive results on EMOTIC dataset and new dataset. 

\textbf{Qualitative results analysis}. In the Figure \ref{fig-Qualitative-Results}, the first column is input image, the second column is depth features. The third column is semantic attention map which represents what our semantic extraction network learned for emotion recognition. The fourth column is top 5 semantic features. The fifth column is top 5 scene features. In the Figure \ref{fig-Qualitative-Results}.(a), the face and body of target person is partially visible. So, our semantic extraction network will pay more attention on the action of baseball player and baseball field. Our model discovered that the baseball player is playing baseball on a baseball field using semantic information and scene characteristics.The depth map reveals that these baseball players are fairly near to one another and they are actively playing the game, which convey  excitement, anticipation, yearning and  engagement. In the Figure \ref{fig-scene-semantic}.(a), the baseball player's hands are up in a defensive posture, as shown by the pose characteristic, and they are acting quite confidently. As a result, our numerous modalities can successfully capture semantic, scene, depth, and posture information to estimate the target person's emotional states.

\section{Conclusion, Limitation and Future work}
\label{sec:conclusion}
In this paper, we propose to use combined scene and semantic features with personal features to reason the emotion states.  Our results show that the combined scene and semantic features can effectively improve the model’s accuracy when the target person’s face and body features are partially visible. We also provide the qualitative result and notice that our network can learn scene and semantic features and pay more attention to the critical parts which influence emotion states of the agent. We hope our study can promote the study in influence of semantic features on emotion recognition. 

\section*{Acknowledgment}
I would like to thank my supervisors for valuable discussion, help and support. This work was partially supported by Australian Government Research Training Program Scholarship.

\end{document}